\pdfoutput=1

\documentclass[11pt]{article}

\usepackage[]{acl}

\usepackage{times}
\usepackage{latexsym}

\usepackage[T1]{fontenc}

\usepackage[utf8]{inputenc}

\usepackage{microtype}

\usepackage{inconsolata}
\usepackage{graphicx}
\usepackage{booktabs}
\usepackage{lipsum}

\title{From Text to Source:\\ Results in Detecting Large Language Model-Generated Content}

\author{Wissam Antoun \quad Benoît Sagot \quad Djamé Seddah \\
     Inria, Paris\\
     \{firstname,lastname\}@inria.fr}

\begin{document}
\maketitle
\begin{abstract}
  The widespread use of Large Language Models (LLMs), celebrated for their ability to generate human-like text, has raised concerns about misinformation and ethical implications.
Addressing these concerns necessitates the development of robust methods to detect and attribute text generated by LLMs.
This paper investigates "Cross-Model Detection," by evaluating whether a classifier trained to distinguish between source LLM-generated and human-written text can also detect text from a target LLM without further training.
The study comprehensively explores various LLM sizes and families, and assesses the impact of conversational fine-tuning techniques, quantization, and watermarking on classifier generalization.
The research also explores Model Attribution, encompassing source model identification, model family, and model size classification, in addition to quantization and watermarking detection.
Our results reveal several key findings: a clear inverse relationship between classifier effectiveness and model size, with larger LLMs being more challenging to detect, especially when the classifier is trained on data from smaller models.
Training on data from similarly sized LLMs can improve detection performance from larger models but may lead to decreased performance when dealing with smaller models.
Additionally, model attribution experiments show promising results in identifying source models and model families, highlighting detectable signatures in LLM-generated text, with particularly remarkable outcomes in watermarking detection, while no detectable signatures of quantization were observed.
Overall, our study contributes valuable insights into the interplay of model size, family, and training data in LLM detection and attribution.
\end{abstract}

\section{Introduction}

Large Language Models (LLMs), characterized by their ability to generate human-like text~\cite{dou-etal-2022-gpt}, have found applications in various domains, including content generation, chatbots, and language translation.
However, as the use of LLMs becomes more widespread, concerns about their misuse, misinformation, and ethical implications have surfaced~\cite{mcguffie2020radicalization,bender2021dangers,chiesurin-etal-2023-dangers}.
One of the ways to address these concerns is with robust methods that are able to detect and attribute text generated by LLMs~\cite{jawahar-etal-2020-automatic}, allowing us to differentiate between human-authored and machine-generated content, identify the source model, or even the model creator.
Such capabilities are crucial for maintaining trust in online communication platforms, content moderation, and ensuring responsible AI deployment.

Our motivation for this research stems from real-life scenarios where we often lack knowledge of the specific model used to generate a piece of text.
These scenarios can be formulated as a "Cross-Model Detection", where we investigate whether a classifier originally trained to distinguish between text generated by one LM and human-written text, can also identify text generated by a different LM without requiring fine-tuning or training on the text it produces.

Our contribution to this area is characterized by the comprehensiveness of our study.
While previous works in the literature have been limited in their exploration of a few model sizes and families, we take a more expansive approach.
We systematically examine a wide range of LLM sizes, spanning from base models to exceptionally large ones, and encompassing diverse model families such as GPT-2, LLaMA, Pythia, OPT and others~\cite{zhao2023survey}.
Additionally, we explore the impact of conversational fine-tuning techniques, including Chat, Instruct~\cite{mishra-etal-2022-cross,wei2022finetuned}, and Reinforcement Learning from Human Feedback (RLHF)~\cite{Christiano2017DeepRL,ziegler2020finetuning}, the impact of model quantization and watermarking of the generated text on the generalization and transferability of the classifier across this wide array of models.

This comprehensive investigation enables us to gain a deeper understanding of the generalization and transferability of the classifier across a diverse array of models, thus eliminating a potential source of bias in our results.
It also allows us to identify how the factors mentioned before (e.g model size, family, conversational finetuning, quantization and watermarking) impact the detection and attribution of generated text. 

Our contributions in this study can be summarized as follows:
\vspace{-0.5em}
\begin{itemize}
    \setlength\itemsep{-0.5em}
    \item A comprehensive investigation into cross-model detection, evaluating the classifier's ability to detect text generated by different LLMs, and in model attribution, encompassing a broad range of sizes and model families.
    \item  We highlight the role of both model size and family in the detection of text generated by Language Model Models (LLMs). We observed an inverse relationship between classifier effectiveness and LLM size. Detecting larger models can be challenging, but training on similarly sized LLMs can improve performance.
    Additionally, our study provides valuable insights into the impact of conversational finetuning, quantization, and watermarking on detection, indicating the robustness of LLM detection in the face of these techniques.
    \item Our experiments in model attribution reveal the potential for identifying the source model of generated text. While human-generated text is distinguishable, confusion primarily occurs between models from the same family or with adjacent sizes. This suggests that LLMs leave distinct signatures, enabling source model identification and model family classification, further enhancing our understanding of how different LLMs generate text.
    
\end{itemize}

In the subsequent sections, we present a summary of relevant related works followed by the details of our methodology, experiments, and results, shedding light on the interplay between model size, family, and training data in the context of LLM detection and attribution in the ever-evolving landscape of Large Language Models.
\section{Related Works}
Detecting AI-generated text is a recent and rapidly growing area of research~\cite{jawahar-etal-2020-automatic}.
Although \citet{sadasivan2023can} demonstrated a theoretical impossibility of distinguishing between human-written and machine-generated when the total variation (TV) norm between the two is low, a more recent study by ~\citet{chakraborty2023possibilities} showed that detection is still possible given enough samples.

Popular methods to detect AI-generated text can be grouped into three categories:
1) Using statistical features of text such as perplexity, n-grams, entropy, etc. \cite{gehrmann-etal-2019-gltr,mitchell2023detectgpt}.
2) Watermarking generated text which was first demonstrated by \citet{atallah2001natural} who embedded a watermark bit in the syntactic structure of the text. More recently, \citet{kirchenbauer2023watermark} used the LLM's output log probability at each generation step to embed a watermark based on ``green/red'' token list where an LLM will have an artificially increased likelihood of selecting tokens from the ``green'' list.
Other work on watermarking include~\cite{fernandez2023bricks,christ2023undetectable}.
3) Classifier-based approaches which use a classifier trained on a dataset containing both human-written and machine-generated text to detect LM-generated text~\cite{zellers2019neuralfakenews,solaiman2019release,uchendu-etal-2020-authorship,fagni2021tweepfake,antoun-etal-2021-aragpt2,guo2023close,mitrovic2023chatgpt}.
This approach is vulnerable against adversarial text mimicking, among others, Wikipedia style and informative~\cite{antoun:hal-04130146}.

We highlight recent work by \citet{mireshghallah2023smaller} that studies cross-model detection and detector transferability by examining the effect of using classifier models other than the generator itself to detect machine-generated text.
The authors studied training LMs from 5 different model families with sizes ranging from 70M to 6.7B and trained the generator LMs to detect machine-generated text generated by other LMs.
They demonstrated that using smaller language models for detection can lead to a higher performance.
Our work differs from \citet{mireshghallah2023smaller} in that we assume we don't have access to the underlying model but only to a set of text generated by the model.
We hence use a separate encoder classifier to detect generated text instead of using the generator.

A similar work to ours was recently published as part of the AuTexTification shared task at IberLEF 2023~\cite{sarvazyan2023overview,Sarvazyan2023supervided}. The shared task offers two subtasks on the detection and attribution of machine-generated text in two languages, English and Spanish, spanning multiple domains. The authors conducted the study on two model families, BLOOM (1.7B, 3B and 7B) and OpenAI's GPT-3 (1B, 3B and 175B).
In addition to this, a new paper, by~\citep{pu2023zero} was recently published, which aligns closely with our work on zero-shot generalization of detector models.
Our study extends the previous work to more model families, sizes and other model specificities.
\section{Methodology}

\paragraph{Cross-Model Detection} Our objective is to evaluate whether a classifier, initially trained to distinguish text produced by a source LLM from human-written text, can also detect text generated by a target LLM.

We conduct a comprehensive evaluation, by using LLMs with a range of sizes (base models to up to very large LLMs) from different families.
We consider a model's family as a proxy for pretraining dataset variation, since apart from slight changes in model architecture, namely positional embeddings type, layer-norm order, or activation type, the only difference between the models from different families is the dataset used for pretraining.

Furthermore, we investigate the effect of Chat, Instruct and Reinforcement Learning from Human Feedback (RLHF) which we refer to as conversational fine-tuning.
To further understand the adaptability and generalization capabilities of the classifier, we also extend our analysis to explore the impact of model quantization and the application of watermarking techniques on the generated text.

This multifaceted approach allows us to evaluate the classifier's effectiveness and transferability across a wide spectrum of LLMs.

\paragraph{Model Attribution} We divide this task into five subtasks:
\vspace{-0.5em}
\begin{itemize}
      \setlength\itemsep{-0.3em}
      \item \textbf{Source Model Identification:} We first investigate the ability to identify the source model of a given piece of text. This investigation should highlight any confusion between the different models, including human-generated text.
      \item \textbf{Model Family Classification:} We classify the source model into its corresponding family.
            This classification helps us understand how good a text can be attributed to a specific model family, and identifies instances where confusion arises between different model families.
            This task is a higher-level generalization of the Source Model Identification task.
      \item \textbf{Model Size Classification:} We examine the ability to determine the model size responsible for generating a given text.
            This experiment aims to determine whether it is feasible to discern whether the text was generated by a small or large LLM.
            This information is valuable for understanding how model size impacts the generated content.
      \item \textbf{Quantization Detection:} We investigate whether a classifier can classify if the source LLM was quantized or using the higher precision FP16 weights.
       \item \textbf{Watermark Detection:} We study if watermarking an LLM output leaves a signature detectable by a classifier model.
\end{itemize}
These investigations collectively contribute to a comprehensive understanding of model attribution in the context of our study.

Our research methodology for investigating cross-model detection and model attribution involves synthetic text generated using Large Language Models (LLMs) selected from diverse families, sizes, and architectures.

\section{Experimental Protocol}
\label{sec:experiments}

\subsection{LLM Choice}
We chose the following model families and sizes for our experiments for a total of 55 models:
\vspace{-0.5em}
\begin{itemize}
    \setlength\itemsep{-0.5em}
    \item \textbf{BLOOM}~\cite{scao2022bloom}: 560M, 1.1B, 1.7B, 3B, 7.1B. 
    \item \textbf{Cereberas-GPT}~\cite{dey2023cerebrasgpt}: 111M, 256M, 1.3B, 2.7B, 6.7B, 13B. 
    \item \textbf{GPT-2}~\cite{radford2019language}: 124M, 355M, 774M, 1.5B. 
    \item \textbf{LLaMA}~\cite{touvron2023llama}: 7B, 13B, 30B, 65B. 
    \item \textbf{LLaMA-v2}~\cite{touvron2023llama2}: 7B, 13B, 70B. 
    \item \textbf{MPT}~\cite{MosaicML2023Introducing}: 7B, 30B. 
    \item \textbf{OPT}~\cite{zhang2022opt}: 125m, 350m, 1.3B, 2.7B, 6.7B, 13B, 30B, 66B. 
    \item \textbf{OpenLLaMA}~\cite{openlm2023openllama}: 3B, 7B, 13B. 
    \item \textbf{OpenLLaMA-v2}~\cite{openlm2023openllama}: 3B, 7B. 
    \item \textbf{Pythia}~\cite{biderman2023pythia}: 70m, 160m, 410m, 1B, 1.4B, 2.8B, 6.9B, 12B. 
\end{itemize}

We select the following conversationally finetuned models to compare with their corresponding foundation models:
\vspace{-0.5em}
\begin{itemize}
    \setlength\itemsep{-0.5em}
    \item \textbf{Falcon-Instruct}~\cite{falcon,refinedweb}: 7B and 40B.
    \item \textbf{Alfred-0723}: 40B, an RLHF finetuned version of Falcon-40B.
    \item \textbf{LLaMA-v2-Chat}~\cite{touvron2023llama2}: 7B, 13B and 70B, an RLHF finetuned version of LLaMA-v2
    \item \textbf{MPT-Chat}~\cite{MosaicML2023Introducing}: 7B, 30B, based on MPT finetuned on a large selection of chat datasets.
    \item \textbf{Vicuna-v1.3}~\cite{zheng2023judging}: 7B, 13B, 33B, based on LLaMA fine-tuned on user-shared conversations.
\end{itemize}

\subsection{Data-Generation}
\label{sec:data.generation}
We generate our data by prompting the different LLMs with the first 10 words of documents sampled from the OpenWebText dataset~\cite{Gokaslan2019OpenWeb}.
For conversational models, in addition to each model's specific prompt, we explicitly instruct the model to continue generation with the following prompt: \textit{``Give the best continuation of the following text:''}, followed by the first 10 words from the sampled document.

We use the HuggingFace Text Generation Inference (TGI) server~\footnote{\url{https://github.com/huggingface/text-generation-inference}} to load all models using up to 4 48GB NVIDIA GPUs for the largest models with float16 precision.
The same set of hyper-parameters is used for all models, a maximum 256 tokens per generation, with beam-search of size 5, repetition penalty of 1.0, temperature of 1.0, top-k of 10, top-p of 0.9 and typical sampling with 0.9.

For model quantization, we chose 4-bit GPTQ quantization~\cite{frantar2023gptq}. As for watermarking, we watermark text using the ``red/green'' token algorithm by \citet{kirchenbauer2023watermark}\footnote{We set gamma to 0.5 and delta to 2.}. We limit our model selection for the quantization and watermarking experiments to the LLaMA and LLaMA-v2 model families.

\subsection{Data Splitting and Filtering}
\label{sec:data.filtering}

We first split our data into 80\% for training and 20\% for validation.
Then we filter each split to remove bad generations, by filtering (i) generations that are too short, (ii) generations that are repetitive and (iii) generations that contain apologies or sentences similar to "As an AI language model".
To ensure a fair comparison between classifier trainings, we sample 800 samples for training and 200 samples for validation from all models.
The \textit{pythia-70m}, \textit{Cereberas-GPT-110m \& 256m} and \textit{OPT-350m} models failed to generate enough valid examples to reach our required sample size and were hence discarded.

For the negative human-generated samples, we sample a new set of texts (800 samples for training and 200 for validation) from the same OpenWebText dataset.

\subsubsection{Cross-Model Detection Training Data}
For each LLM we merge its own train and test sets with the negative examples sets for a total of 1600 training samples and 400 validation samples.
To quantify classifier performance, and following~\cite{sadasivan2023can,chakraborty-2023-rgat}, we utilize the Area Under the Receiver Operating Characteristic Curve (AUC score).
The AUC score provides a robust measure of the classifier's ability to distinguish between different models, taking into account both true positive and false positive rates.

\subsubsection{Model Attribution Training Data}
\label{sec:model_attr}

We conduct three distinct investigations as mentioned earlier. We use F1 score to evaluate classifier performance in all three tasks due to the imbalanced nature of our data.

\paragraph{Source Model Identification}
This task involves classifying the text into one of the 50 LLMs used in our experiments, spanning various families and sizes.
We also include a class for human written text for a total of 51 classes.

\paragraph{Family Classification}
We group models into 12 classes including one for human written text, and then sub-sample the data to have a balanced 1600 training samples and 400 validation samples for each class.

\paragraph{Model Size Identification}
We bin the models into 6 bins: <1B, 1-5B, 5-10B, 10-20B, 20-50B, >50B.
Refer to Table~\ref{tab:size_distribution} for the class distribution.

\begin{figure*}[!ht]
    \includegraphics[width=\textwidth]{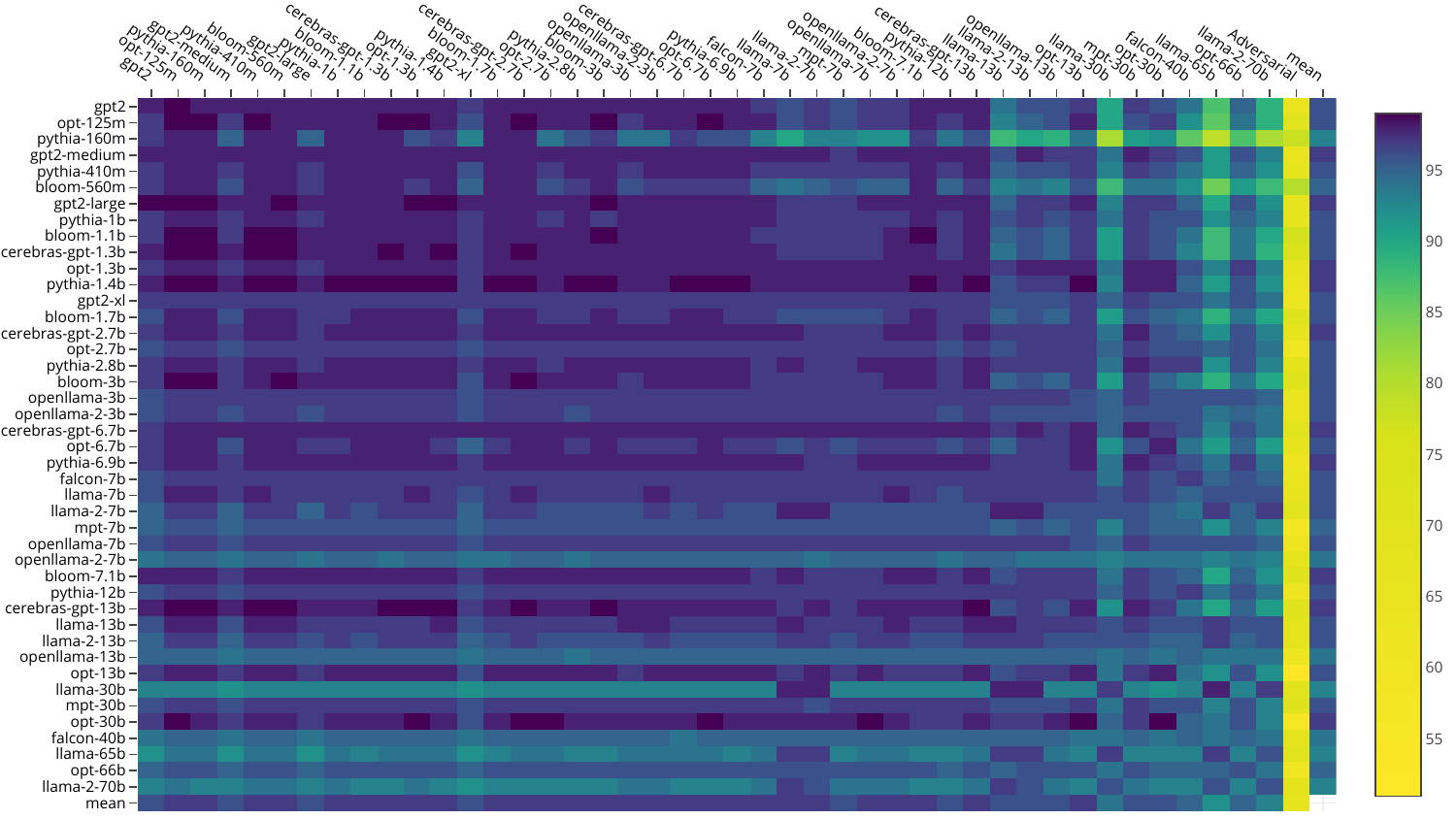}
    \caption{5-seed averaged AUC scores for a classifier trained on text from a source model (\textit{Side axis}) and tested on text from a target model (\textit{Top axis}).\textit{ We provide the full results table with values in Appendix~\ref{appendix:full_table}.}}
    \label{fig:res.cross.det}
\end{figure*}

\begin{table}[!ht]
    \caption{Model Size Distribution}
    \centering\small
    \begin{tabular}{rrr}
        \toprule
        Size Bin & Train & Test \\ \midrule
        <1B      & 9600  & 2400 \\
        1-5B     & 11200 & 2800 \\
        5-10B    & 12000 & 3000 \\
        10-20B   & 7200  & 1800 \\
        20-50B   & 6400  & 1600 \\
        >50B     & 3200  & 800  \\
        \bottomrule
    \end{tabular}
    \label{tab:size_distribution}
\end{table}

\paragraph{Quantization and Watermarking Detection}
For both experiments, we have 11k balanced training samples and 2800 testing samples.

\subsection{Classifier}
\label{sec:classifier}
Finetuning an encoder model is a popular approach for AI-generated text detection, and has been used widely~\cite{guo2023close,sarvazyan2023overview,crothers2023machine}.
Hence, our classifier of choice for all experiments uses the transformer encoder architecture namely DeBERTaV3-base~\cite{he2023debertav,he2021deberta}.
All models are trained with a batch size of 32, learning rate of 2e-5 for 5 epochs.
The classification experiments were conducted using five different random seeds, and the resultant scores were averaged to enhance the robustness of our findings, as this approach helps mitigate the potential impact of seed-specific variations on the results.

\begin{figure*}[t!]
    \includegraphics[width=\textwidth]{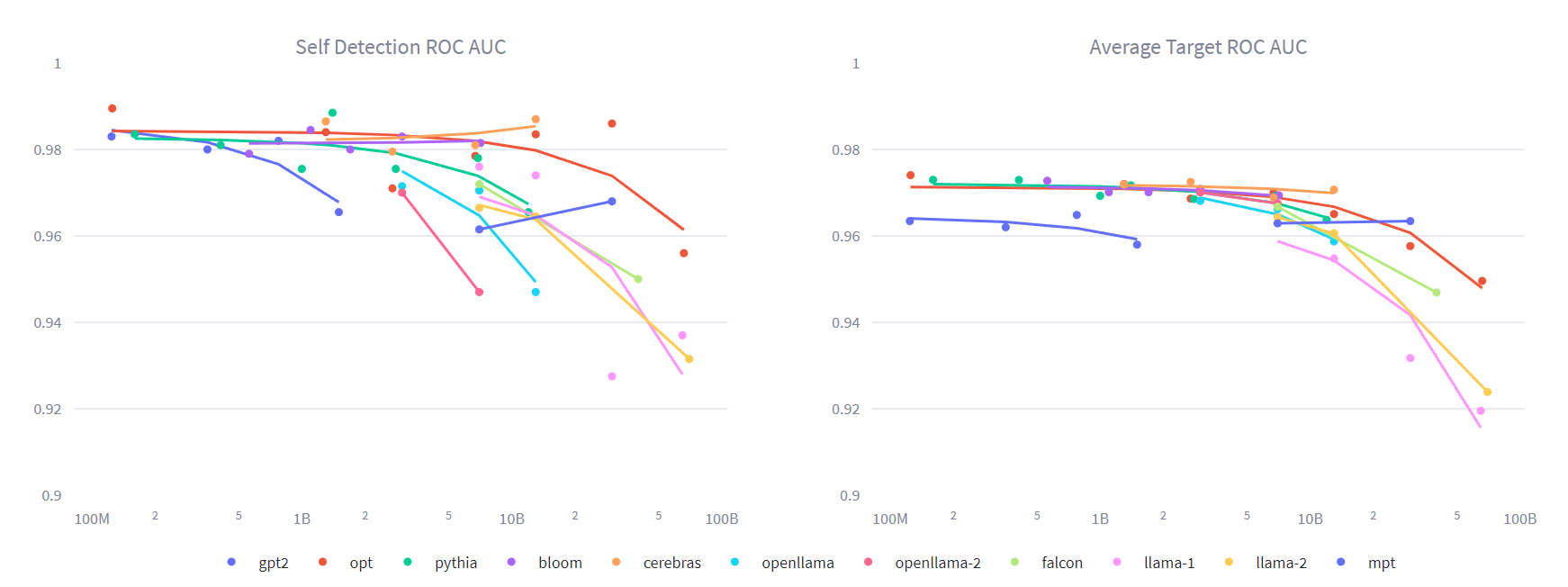}
    \caption{Average target AUC scores vs model size. OLS Trend lines are drawn for each set of model family}
    \label{fig:res.self.avgperf.vs.size}
\end{figure*}

\section{Results}
\subsection{Cross-Model Detection Results}
\label{sec:res.cross.detect}
Figure~\ref{fig:res.cross.det} presents a heatmap of the AUC scores for the cross-model detection experiments.
The side axis represents the classifier's source model, and the top axis represents the classifier's target model.
We sort the models by their size (from left to right, top to bottom).
From Figure~\ref{fig:res.cross.det}, we observe several interesting patterns in the cross-model detection results:

\paragraph{Model Size Influence}
In general, our findings suggest a clear inverse relationship between the classifier's effectiveness and the size of the test models.
The pattern is showcased better in Figure~\ref{fig:res.self.avgperf.vs.size}, which plots the self-detection and average AUC scores trend lines against the model size.
This pattern indicates that larger LLMs tend to pose a greater challenge for the classifier, particularly when the classifier is trained on data from a smaller source model.
Notably, the detection performance on very large Language Models (LMs) tends to improve when the model is trained on data sourced from similarly sized large LMs.
However, it is essential to highlight the trade-off that training only on very large LMs leads to, results in decreased performance in detecting smaller-sized models.

\paragraph{Model Family Influence}
We observe that performance on detecting GPT2 and LLaMA generated text tends to be slightly lower than other model families \textit{(Refer to the corresponding heatmap columns and their means in Figure~\ref{fig:res.cross.det} to observe the corresponding data patterns)}.
This pattern suggests that the two model families are harder to detect relative to their similar-sized counterparts due to their superior language modeling capabilities and hence ``closer'' to human written text.
We can also observe that the performance of a classifier trained on text sourced from \textit{pythia-160m} and \textit{opt-2.7b} tends worse overall, while a classifier trained on text sourced from \textit{Cereberas-GPT-6.7B} is performing better than similarly sized models \textit{(Refer to the corresponding heatmap rows in Figure~\ref{fig:res.cross.det})}.
\textbf{The lack of a discernible pattern in the cross-model detection performance across different model families may be attributed to the extensive overlap in their pretraining data}, with a predominant reliance on ThePile~\cite{gao2020pile} dataset or its subsets across most models, supplemented by Common Crawl as the primary data source. Consequently, the primary distinguishing factor among these models lies in their respective data cleaning pipelines.

\begin{figure}[ht]
    \includegraphics[width=\columnwidth]{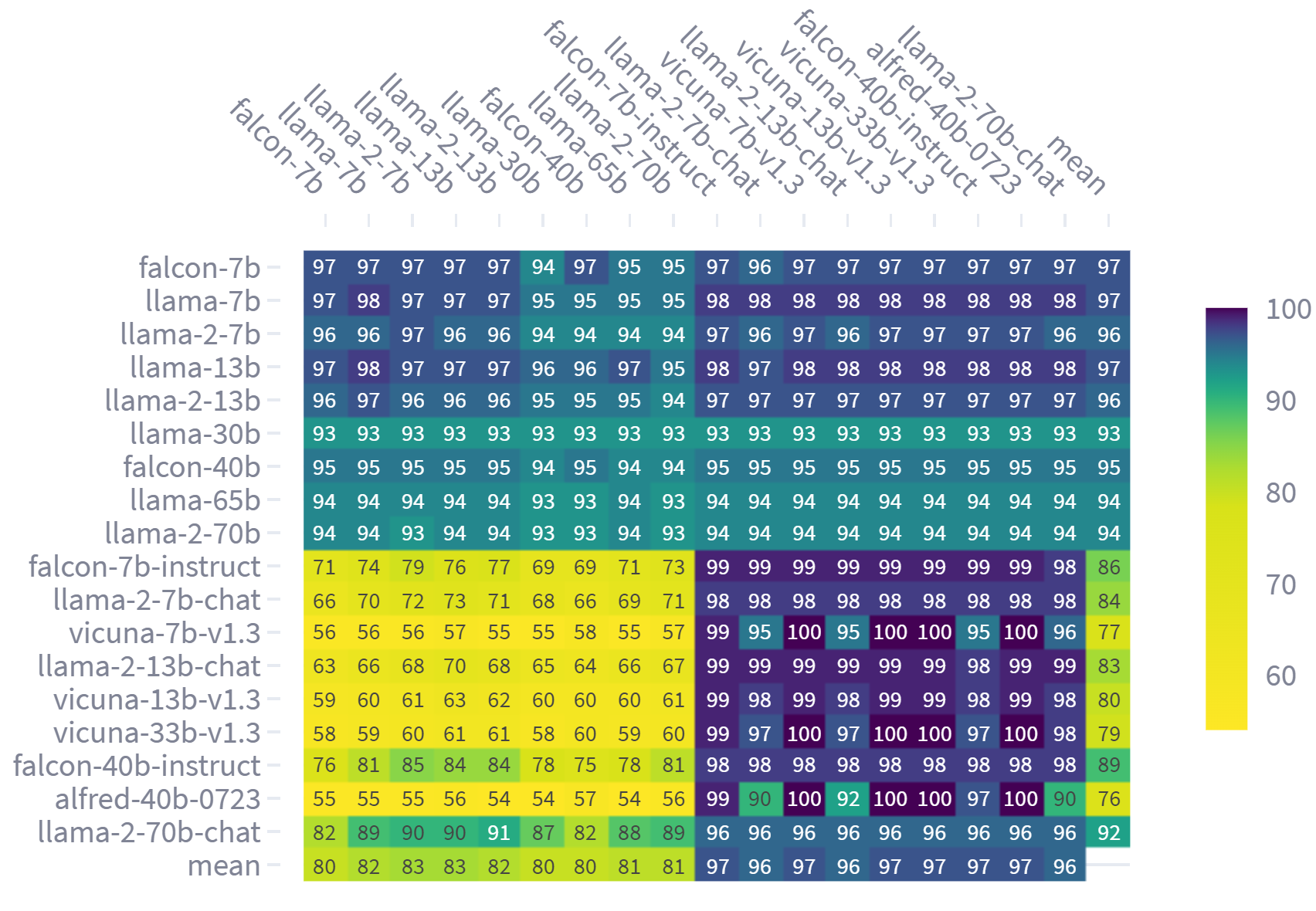}
    \caption{Conversational models cross-model detection with their foundation LLM (5-seed averaged AUCROC score).}
    \label{fig:res.chat.cross.det}
\end{figure}

\paragraph{Influence of Conversational Finetuning}
Our experiments reveal a clear pattern in the cross-model detection results, as shown in Figure~\ref{fig:res.chat.cross.det}.
Specifically, a classifier trained on text generated by chat models exhibits limited capability in detecting normal language models (LMs).
However, it demonstrates improved performance when tasked with detecting other chat models.
Notably, when trained on \textit{LLaMA-2-70b-chat} data, the classifier achieves the highest scores, albeit with a slight decline in detection accuracy when tested on chat models.
This observation suggests that the \textit{LLaMA-2-70b-chat} model effectively follows instructions to continue input text.
Surprisingly, training the classifier on vanilla LM output also yields commendable results in detecting these distinct model categories.
These findings underscore the nuanced relationship between chat models and traditional language models in the context of detection.

\begin{figure*}[!ht]
    \includegraphics[trim=0 60 10 0,clip,width=\textwidth]{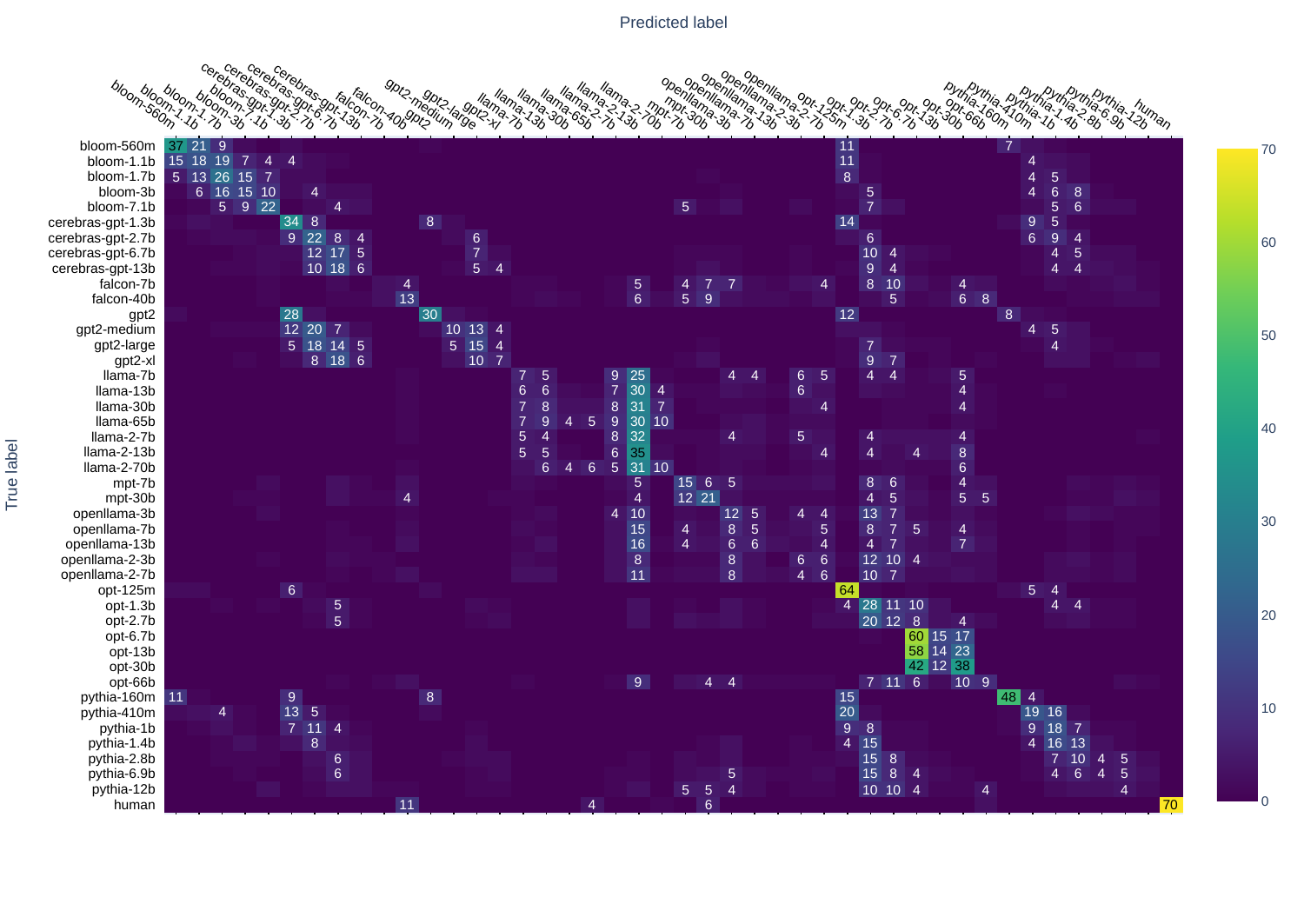}
    \caption{Normalized confusion matrix for Source Model Identification. 5-seed averaged and normalized by the predicted class support. \textit{For clarity purposes, we hide labels on values less than 4.}}
    \label{fig:res.attribution}
\end{figure*}

\begin{figure}[ht]
    \includegraphics[width=\columnwidth]{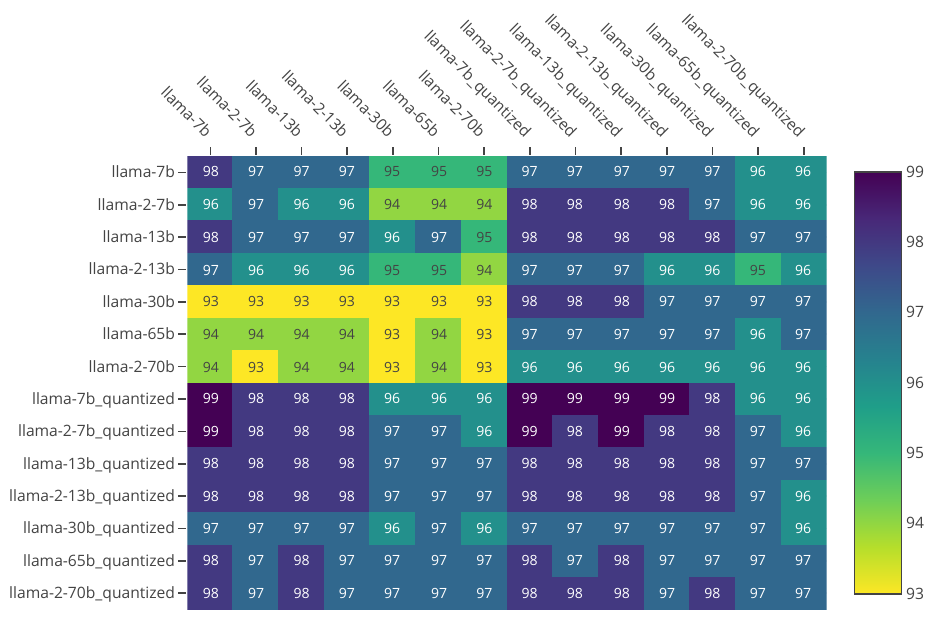}
    \caption{Quantized models cross-model detection (5-seed averaged AUCROC score).}
    \label{fig:res.quant.cross.det}
\end{figure}

\begin{figure}[ht]
    \includegraphics[width=\columnwidth]{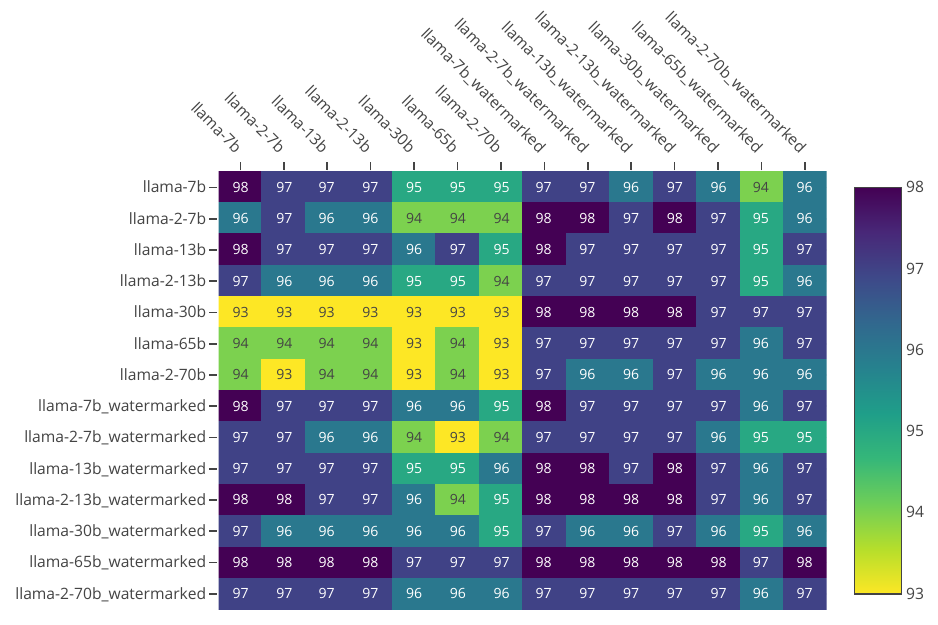}
    \caption{Watermarked models cross-model detection (5-seed averaged AUCROC score).}
    \label{fig:res.watermrk.cross.det}
\end{figure}

\paragraph{Influence of Quantization and Watermarking}
The results in Figure~\ref{fig:res.quant.cross.det} show a weak pattern, wherein a classifier trained using text data from a quantized model exhibits a slightly higher degree of transferability when compared to training with conventional language models.
A similar trend is seen in Figure~\ref{fig:res.watermrk.cross.det} with the watermarking experiment.
It is worth noting that the underlying reasons for this phenomenon remain unclear, and further investigation remains a potential direction for future research.

\subsection{Model Attribution Results}
\label{sec:res.attribution}

\paragraph{Source Model Identification}
\label{sec:res.source.classification}
In the Model Attribution experiments, our objective was to investigate the ability of our classifier to identify the source model of generated text accurately.
Figure~\ref{fig:res.attribution} displays the confusion matrix for the Model Attribution experiments, where rows represent the true source models, and columns represent the predicted source models.
We can draw the following conclusions from our results:
\vspace{-0.5em}
\begin{itemize}
    \setlength\itemsep{-0.5em}
    \item Human-generated text proved to be the most easily distinguishable source, as it exhibited minimal confusion, primarily with a few 30B+ Large Language Models (LLMs).
    \item The majority of confusions occurred between models from the same family. We also notice that within a model family, the confusions tend to happen between models with adjacent sizes.
    \item An interesting case was the confusion between GPT-2 models and Cereberas-GPT models. It's worth noting that both models share the same GPT-2 architecture but differ in their pretraining data, with Cereberas-GPT being trained on ThePile, which includes an open replication of the OpenWebText dataset.
\end{itemize}
Overall, our classifier achieved an F1-score of 17.7\% across 44 distinct labels, indicating that LLMs leave detectable signatures, thus enabling source model identification.

\begin{figure}[t!]
    \includegraphics[trim=0 60 10 0,clip,width=\columnwidth]{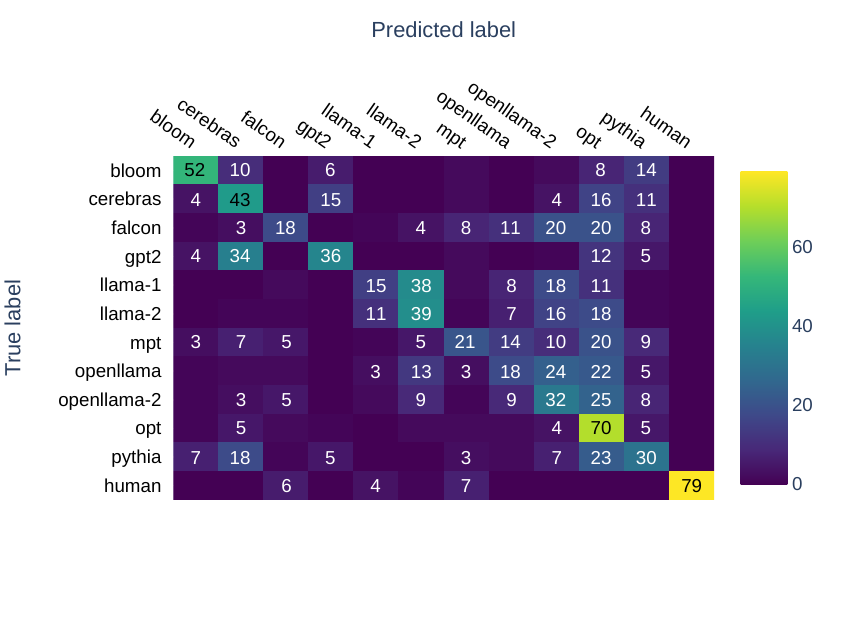}
    \caption{Normalized confusion matrix for model family classification. 5-seed averaged and normalized by the predicted class support. \textit{For clarity purposes, we hide labels on values less than 4.}}
    \label{fig:res.family.classification}
\end{figure}

\paragraph{Model Family Classification}
\label{sec:res.family.classification}
In the Model Family Classification experiments, our primary aim was to evaluate the classifier's efficacy in identifying the model family responsible for generating a given text. This assessment allows us to determine if distinct signatures emerge from the diverse pretraining data and architectural choices utilized by different language models.
Figure~\ref{fig:res.family.classification} provides an overview of the Model Family Classification results. Notably, we observe that human-generated text exhibits the highest distinguishability from other model families, followed by the OPT model family. It's worth noting that this distinction might be partly influenced by the subpar generation quality of the OPT-125m model, which stands out and can be easily identified among the models as seen in Section~\ref{sec:res.source.classification}.
Furthermore, we notice a consistent confusion pattern between GPT-2 models and Cereberas-GPT models. These two model families, sharing the GPT-2 architecture but differing in their pretraining data sources, appear to exhibit a higher degree of similarity in their generated text, leading to increased misclassifications.
The overall F1-score across 12 distinct model family labels was 37\%, underscoring the potential for detecting model family signatures.

\begin{figure}[t]
    \includegraphics[trim=0 60 10 0,clip,width=\columnwidth]{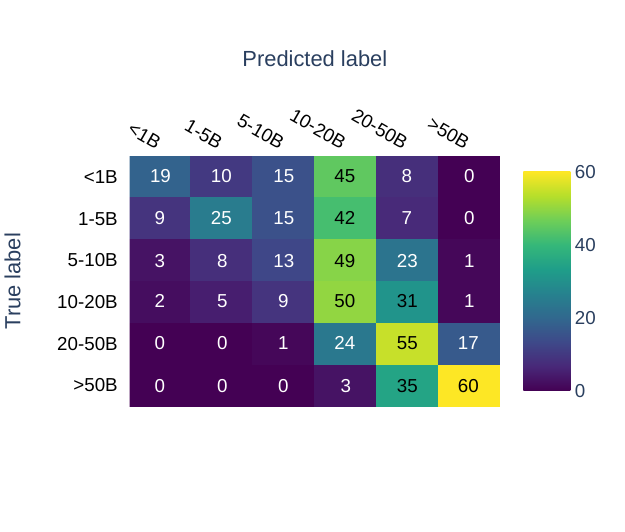}
    \caption{Normalized confusion matrix for model size classification (5-seed averaged).}
    \label{fig:res.size.classification}
\end{figure}

\paragraph{Model Size Classification}
\label{sec:res.size.classification}
In the Model Size Classification experiments, we aimed to assess the classifier's ability to determine the size category of the model responsible for generating a given text.
This evaluation allows us to discern whether the differences in model sizes translate into detectable signatures in the generated text.
As depicted in Figure~\ref{fig:res.size.classification}, the results of the Model Size Classification experiment reveal a discernible pattern.
Larger models consistently exhibit the least amount of confusion, while models with sizes that are closely related tend to be more frequently misclassified.
An interesting exception to this pattern is observed in the case of the 10-20B model, where the classifier tends to confuse other smaller models with it.
In summary, the classifier achieves an overall F1-score of 38\% across six distinct model size categories.

\paragraph{Quantization Detection}
In this experimental evaluation, our classifier fails to detect patterns or distinctions between quantized and unquantized models, yielding an accuracy of $54.5\%\pm0.9$.
These findings provide additional evidence supporting the effectiveness of GPTQ as a quantization method, which appears to leave no discernible traces or artifacts within the generated text.

\paragraph{Watermark Detection}
This experiment demonstrates our ability to effectively train a classifier to identify watermarked text, achieving an accuracy of $82.3\%\pm2.1$. This finding has significant implications, as it implies that \textbf{watermark signatures can be identified and disclosed through the use of an encoder classifier}, without requiring access to the source model's log probabilities or any prior knowledge thereof.
\section{Discussion}
\label{sec:discussion}

The experiments and results presented in this study provide valuable insights into the challenges and nuances of detecting and attributing text generated by different LLMs.

In the cross-model detection experiments, we observed a clear inverse relationship between the effectiveness of the classifier and the size of the test models.
Larger LLMs tend to be more challenging to detect, especially when the classifier is trained on data from smaller models.
However, training on similarly sized LLMs can improve detection performance on larger models, although it may lead to decreased performance on smaller models.
Interestingly, the performance varied across LLM families, with GPT2 and LLaMA-generated text proving harder to detect due to their advanced language modeling capabilities.
These findings emphasize the importance of considering both model size and family when developing detection strategies.
In addition to the observations made in the cross-model detection experiments, we also conducted experiments to assess the influence of conversational models, shedding light on the relationship between chat models and traditional language models in the context of detection.

In the Model Attribution experiments, our classifier demonstrated the ability to identify the source model of generated text to a certain extent.
Human-generated text was the most distinguishable, Figure~\ref{fig:res.attribution} and \ref{fig:res.family.classification}, while confusion mainly occurred between models from the same family and between models with adjacent sizes.
Furthermore, in Model Family Classification, the classifier showed promise in identifying the model family responsible for generating text, highlighting the potential for detecting distinct signatures arising from diverse pretraining data and architectural choices.
This indicates that LLMs leave detectable signatures, enabling source model identification, and model family classification.
In Model Size Classification, we observed that larger models were less frequently misclassified, emphasizing the influence of model size on detection.
More importantly, our classifier was able to detect watermarking artifacts with only access to the generated text. While the failure to detect quantization of the source model further affirms the negligible loss of perplexity when using GPTQ as shown by~\citet{frantar2023gptq}.

Building upon the findings of~\citet{antoun:hal-04130146} which demonstrated the challenging nature of identifying adversarial text composed in an academic, pedagogic, or encyclopedic style for state-of-the-art classifiers trained on a mixture of text generated by LLMs and human content, we also investigated how the detection of adversarial content text could influence the trends we exposed earlier.
As shown in Figure~\ref{fig:res.cross.det}, in the Adversarial column, the results are massively inferior to the ones reported in our main experimental setting.
The inherent out-of-domain distribution of this content\footnote{ We translated the original data set from French to English using google translate. As shown by~\citet{antoun:hal-04130146}, translated text from English to French has no effect in the detectability of generated content. We believe this holds in the French to English direction.} compared to our main experiment setting may have indeed contributed significantly to this performance degradation.

This observation suggests that these detectors are likely taking advantage of relevant textual features to distinguish between automatically generated text of lower quality and human-produced content.
However, it should be acknowledged that the results exhibit variability across models, with models of similar size encountering difficulties in this task, while larger model-trained classifiers also face challenges in this specific context.

Further work is required to investigate the precise factors at play in this scenario.
Our key takeaway is that our study was conducted within a controlled environment, aiming to single-out variable influences.
Therefore, the level of performance we demonstrated should not be interpreted as indicative of real-world expectations for this task.
We envision for the future that a detectability score can be used as a proxy or as an evaluation of a model's quality.
Overall, our results underscore the complex interplay between model size, family, and training data in the context of LLM detection and attribution.
Moreover, to enhance transparency, we provide all our experiments results in an interactive online repository \url{https://huggingface.co/spaces/wissamantoun/LLM_Detection_Attribution}.

\section{Limitations}

It is crucial to acknowledge several limitations that warrant further investigation.
We did not explore the impact of various sampling strategies or parameters, such as temperature, which could influence the results.
Secondly, our study solely focused on openly available models, excluding black box models accessible only through APIs.
Lastly, we constrained our classification technique to fine-tuning a single model, potentially overlooking alternative approaches.

\section*{Acknowledgements}
This work was partly funded by Benoît Sagot's chair in the PRAIRIE institute funded by the French national research agency (ANR as part of the ``Investissements d’avenir'' programme under the reference \mbox{ANR-19-P3IA-0001}. This work  also received
funding from the European Union’s Horizon 2020 research and innovation program under grant agreement No. 101021607.
The authors are grateful to the OPAL infrastructure from Université Côte d'Azur for providing resources and support.

We would also like to thank Francis Kulumba, Arij Riabi, and Roman Castagné for the productive discussions.

\bibliography{custom}

\begin{thebibliography}{49}
\expandafter\ifx\csname natexlab\endcsname\relax\def\natexlab#1{#1}\fi

\bibitem[{Almazrouei et~al.(2023)Almazrouei, Alobeidli, Alshamsi, Cappelli, Cojocaru, Alhammadi, Daniele, Heslow, Launay, Malartic, Noune, Pannier, and Penedo}]{falcon}
Ebtesam Almazrouei, Hamza Alobeidli, Abdulaziz Alshamsi, Alessandro Cappelli, Ruxandra Cojocaru, Maitha Alhammadi, Mazzotta Daniele, Daniel Heslow, Julien Launay, Quentin Malartic, Badreddine Noune, Baptiste Pannier, and Guilherme Penedo. 2023.
\newblock The falcon series of language models: Towards open frontier models.

\bibitem[{Antoun et~al.(2021)Antoun, Baly, and Hajj}]{antoun-etal-2021-aragpt2}
Wissam Antoun, Fady Baly, and Hazem Hajj. 2021.
\newblock \href {https://aclanthology.org/2021.wanlp-1.21} {{A}ra{GPT}2: Pre-trained transformer for {A}rabic language generation}.
\newblock In \emph{Proceedings of the Sixth Arabic Natural Language Processing Workshop}, pages 196--207, Kyiv, Ukraine (Virtual). Association for Computational Linguistics.

\bibitem[{Antoun et~al.(2023)Antoun, Mouilleron, Sagot, and Seddah}]{antoun:hal-04130146}
Wissam Antoun, Virginie Mouilleron, Beno{\^i}t Sagot, and Djam{\'e} Seddah. 2023.
\newblock \href {https://hal.science/hal-04130146} {{Towards a Robust Detection of Language Model-Generated Text: Is ChatGPT that easy to detect?}}
\newblock In \emph{{18e Conf{\'e}rence en Recherche d'Information et Applications -- 16e Rencontres Jeunes Chercheurs en RI -- 30e Conf{\'e}rence sur le Traitement Automatique des Langues Naturelles -- 25e Rencontre des {\'E}tudiants Chercheurs en Informatique pour le Traitement Automatique des Langues}}, pages 14--27, Paris, France. {ATALA}.

\bibitem[{Atallah et~al.(2001)Atallah, Raskin, Crogan, Hempelmann, Kerschbaum, Mohamed, and Naik}]{atallah2001natural}
Mikhail~J Atallah, Victor Raskin, Michael Crogan, Christian Hempelmann, Florian Kerschbaum, Dina Mohamed, and Sanket Naik. 2001.
\newblock Natural language watermarking: Design, analysis, and a proof-of-concept implementation.
\newblock In \emph{Information Hiding: 4th International Workshop, IH 2001 Pittsburgh, PA, USA, April 25--27, 2001 Proceedings 4}, pages 185--200. Springer.

\bibitem[{Bender et~al.(2021)Bender, Gebru, McMillan-Major, and Shmitchell}]{bender2021dangers}
Emily~M. Bender, Timnit Gebru, Angelina McMillan-Major, and Shmargaret Shmitchell. 2021.
\newblock \href {https://doi.org/10.1145/3442188.3445922} {On the dangers of stochastic parrots: Can language models be too big?}
\newblock In \emph{Proceedings of the 2021 ACM Conference on Fairness, Accountability, and Transparency}, FAccT '21, page 610–623, New York, NY, USA. Association for Computing Machinery.

\bibitem[{Biderman et~al.(2023)Biderman, Schoelkopf, Anthony, Bradley, O'Brien, Hallahan, Khan, Purohit, Prashanth, Raff, Skowron, Sutawika, and van~der Wal}]{biderman2023pythia}
Stella Biderman, Hailey Schoelkopf, Quentin Anthony, Herbie Bradley, Kyle O'Brien, Eric Hallahan, Mohammad~Aflah Khan, Shivanshu Purohit, USVSN~Sai Prashanth, Edward Raff, Aviya Skowron, Lintang Sutawika, and Oskar van~der Wal. 2023.
\newblock \href {http://arxiv.org/abs/2304.01373} {Pythia: A suite for analyzing large language models across training and scaling}.

\bibitem[{Chakraborty(2023)}]{chakraborty-2023-rgat}
Abir Chakraborty. 2023.
\newblock \href {https://doi.org/10.18653/v1/2023.semeval-1.23} {{RGAT} at {S}em{E}val-2023 task 2: Named entity recognition using graph attention network}.
\newblock In \emph{Proceedings of the 17th International Workshop on Semantic Evaluation (SemEval-2023)}, pages 163--170, Toronto, Canada. Association for Computational Linguistics.

\bibitem[{Chakraborty et~al.(2023)Chakraborty, Bedi, Zhu, An, Manocha, and Huang}]{chakraborty2023possibilities}
Souradip Chakraborty, Amrit~Singh Bedi, Sicheng Zhu, Bang An, Dinesh Manocha, and Furong Huang. 2023.
\newblock On the possibilities of ai-generated text detection.
\newblock \emph{arXiv preprint arXiv:2304.04736}.

\bibitem[{Chiesurin et~al.(2023)Chiesurin, Dimakopoulos, Sobrevilla~Cabezudo, Eshghi, Papaioannou, Rieser, and Konstas}]{chiesurin-etal-2023-dangers}
Sabrina Chiesurin, Dimitris Dimakopoulos, Marco~Antonio Sobrevilla~Cabezudo, Arash Eshghi, Ioannis Papaioannou, Verena Rieser, and Ioannis Konstas. 2023.
\newblock \href {https://doi.org/10.18653/v1/2023.findings-acl.60} {The dangers of trusting stochastic parrots: Faithfulness and trust in open-domain conversational question answering}.
\newblock In \emph{Findings of the Association for Computational Linguistics: ACL 2023}, pages 947--959, Toronto, Canada. Association for Computational Linguistics.

\bibitem[{Christ et~al.(2023)Christ, Gunn, and Zamir}]{christ2023undetectable}
Miranda Christ, Sam Gunn, and Or~Zamir. 2023.
\newblock Undetectable watermarks for language models.
\newblock \emph{arXiv preprint arXiv:2306.09194}.

\bibitem[{Christiano et~al.(2017)Christiano, Leike, Brown, Martic, Legg, and Amodei}]{Christiano2017DeepRL}
Paul~F Christiano, Jan Leike, Tom Brown, Miljan Martic, Shane Legg, and Dario Amodei. 2017.
\newblock \href {https://proceedings.neurips.cc/paper_files/paper/2017/file/d5e2c0adad503c91f91df240d0cd4e49-Paper.pdf} {Deep reinforcement learning from human preferences}.
\newblock In \emph{Advances in Neural Information Processing Systems}, volume~30. Curran Associates, Inc.

\bibitem[{Crothers et~al.(2023)Crothers, Japkowicz, and Viktor}]{crothers2023machine}
Evan Crothers, Nathalie Japkowicz, and Herna~L Viktor. 2023.
\newblock Machine-generated text: A comprehensive survey of threat models and detection methods.
\newblock \emph{IEEE Access}.

\bibitem[{Dey et~al.(2023)Dey, Gosal, Zhiming, Chen, Khachane, Marshall, Pathria, Tom, and Hestness}]{dey2023cerebrasgpt}
Nolan Dey, Gurpreet Gosal, Zhiming, Chen, Hemant Khachane, William Marshall, Ribhu Pathria, Marvin Tom, and Joel Hestness. 2023.
\newblock \href {http://arxiv.org/abs/2304.03208} {Cerebras-gpt: Open compute-optimal language models trained on the cerebras wafer-scale cluster}.

\bibitem[{Dou et~al.(2022)Dou, Forbes, Koncel-Kedziorski, Smith, and Choi}]{dou-etal-2022-gpt}
Yao Dou, Maxwell Forbes, Rik Koncel-Kedziorski, Noah~A. Smith, and Yejin Choi. 2022.
\newblock \href {https://doi.org/10.18653/v1/2022.acl-long.501} {Is {GPT}-3 text indistinguishable from human text? scarecrow: A framework for scrutinizing machine text}.
\newblock In \emph{Proceedings of the 60th Annual Meeting of the Association for Computational Linguistics (Volume 1: Long Papers)}, pages 7250--7274, Dublin, Ireland. Association for Computational Linguistics.

\bibitem[{Fagni et~al.(2021)Fagni, Falchi, Gambini, Martella, and Tesconi}]{fagni2021tweepfake}
Tiziano Fagni, Fabrizio Falchi, Margherita Gambini, Antonio Martella, and Maurizio Tesconi. 2021.
\newblock Tweepfake: About detecting deepfake tweets.
\newblock \emph{Plos one}, 16(5):e0251415.

\bibitem[{Fernandez et~al.(2023)Fernandez, Chaffin, Tit, Chappelier, and Furon}]{fernandez2023bricks}
Pierre Fernandez, Antoine Chaffin, Karim Tit, Vivien Chappelier, and Teddy Furon. 2023.
\newblock \href {http://arxiv.org/abs/2308.00113} {Three bricks to consolidate watermarks for large language models}.

\bibitem[{Frantar et~al.(2023)Frantar, Ashkboos, Hoefler, and Alistarh}]{frantar2023gptq}
Elias Frantar, Saleh Ashkboos, Torsten Hoefler, and Dan Alistarh. 2023.
\newblock \href {http://arxiv.org/abs/2210.17323} {Gptq: Accurate post-training quantization for generative pre-trained transformers}.

\bibitem[{Gao et~al.(2020)Gao, Biderman, Black, Golding, Hoppe, Foster, Phang, He, Thite, Nabeshima, Presser, and Leahy}]{gao2020pile}
Leo Gao, Stella Biderman, Sid Black, Laurence Golding, Travis Hoppe, Charles Foster, Jason Phang, Horace He, Anish Thite, Noa Nabeshima, Shawn Presser, and Connor Leahy. 2020.
\newblock \href {http://arxiv.org/abs/2101.00027} {The pile: An 800gb dataset of diverse text for language modeling}.

\bibitem[{Gehrmann et~al.(2019)Gehrmann, Strobelt, and Rush}]{gehrmann-etal-2019-gltr}
Sebastian Gehrmann, Hendrik Strobelt, and Alexander Rush. 2019.
\newblock \href {https://doi.org/10.18653/v1/P19-3019} {{GLTR}: Statistical detection and visualization of generated text}.
\newblock In \emph{Proceedings of the 57th Annual Meeting of the Association for Computational Linguistics: System Demonstrations}, pages 111--116, Florence, Italy. Association for Computational Linguistics.

\bibitem[{Geng and Liu(2023)}]{openlm2023openllama}
Xinyang Geng and Hao Liu. 2023.
\newblock \href {https://github.com/openlm-research/open_llama} {Openllama: An open reproduction of llama}.

\bibitem[{Gokaslan et~al.(2019)Gokaslan, Cohen, Pavlick, and Tellex}]{Gokaslan2019OpenWeb}
Aaron Gokaslan, Vanya Cohen, Ellie Pavlick, and Stefanie Tellex. 2019.
\newblock Openwebtext corpus.
\newblock \url{http://Skylion007.github.io/OpenWebTextCorpus}.

\bibitem[{Guo et~al.(2023)Guo, Zhang, Wang, Jiang, Nie, Ding, Yue, and Wu}]{guo2023close}
Biyang Guo, Xin Zhang, Ziyuan Wang, Minqi Jiang, Jinran Nie, Yuxuan Ding, Jianwei Yue, and Yupeng Wu. 2023.
\newblock How close is chatgpt to human experts? comparison corpus, evaluation, and detection.
\newblock \emph{arXiv preprint arXiv:2301.07597}.

\bibitem[{He et~al.(2023)He, Gao, and Chen}]{he2023debertav}
Pengcheng He, Jianfeng Gao, and Weizhu Chen. 2023.
\newblock \href {https://openreview.net/forum?id=sE7-XhLxHA} {De{BERT}av3: Improving de{BERT}a using {ELECTRA}-style pre-training with gradient-disentangled embedding sharing}.
\newblock In \emph{The Eleventh International Conference on Learning Representations}.

\bibitem[{He et~al.(2021)He, Liu, Gao, and Chen}]{he2021deberta}
Pengcheng He, Xiaodong Liu, Jianfeng Gao, and Weizhu Chen. 2021.
\newblock \href {https://openreview.net/forum?id=XPZIaotutsD} {{\{}DEBERTA{\}}: {\{}DECODING{\}}-{\{}enhanced{\}} {\{}bert{\}} {\{}with{\}} {\{}disentangled{\}} {\{}attention{\}}}.
\newblock In \emph{International Conference on Learning Representations}.

\bibitem[{Jawahar et~al.(2020)Jawahar, Abdul-Mageed, and Lakshmanan}]{jawahar-etal-2020-automatic}
Ganesh Jawahar, Muhammad Abdul-Mageed, and Laks Lakshmanan, V.S. 2020.
\newblock \href {https://doi.org/10.18653/v1/2020.coling-main.208} {Automatic detection of machine generated text: A critical survey}.
\newblock In \emph{Proceedings of the 28th International Conference on Computational Linguistics}, pages 2296--2309, Barcelona, Spain (Online). International Committee on Computational Linguistics.

\bibitem[{Kirchenbauer et~al.(2023)Kirchenbauer, Geiping, Wen, Katz, Miers, and Goldstein}]{kirchenbauer2023watermark}
John Kirchenbauer, Jonas Geiping, Yuxin Wen, Jonathan Katz, Ian Miers, and Tom Goldstein. 2023.
\newblock A watermark for large language models.
\newblock \emph{arXiv preprint arXiv:2301.10226}.

\bibitem[{McGuffie and Newhouse(2020)}]{mcguffie2020radicalization}
Kris McGuffie and Alex Newhouse. 2020.
\newblock The radicalization risks of gpt-3 and advanced neural language models.
\newblock \emph{arXiv preprint arXiv:2009.06807}.

\bibitem[{Mireshghallah et~al.(2023)Mireshghallah, Mattern, Gao, Shokri, and Berg-Kirkpatrick}]{mireshghallah2023smaller}
Fatemehsadat Mireshghallah, Justus Mattern, Sicun Gao, Reza Shokri, and Taylor Berg-Kirkpatrick. 2023.
\newblock \href {http://arxiv.org/abs/2305.09859} {Smaller language models are better black-box machine-generated text detectors}.

\bibitem[{Mishra et~al.(2022)Mishra, Khashabi, Baral, and Hajishirzi}]{mishra-etal-2022-cross}
Swaroop Mishra, Daniel Khashabi, Chitta Baral, and Hannaneh Hajishirzi. 2022.
\newblock \href {https://doi.org/10.18653/v1/2022.acl-long.244} {Cross-task generalization via natural language crowdsourcing instructions}.
\newblock In \emph{Proceedings of the 60th Annual Meeting of the Association for Computational Linguistics (Volume 1: Long Papers)}, pages 3470--3487, Dublin, Ireland. Association for Computational Linguistics.

\bibitem[{Mitchell et~al.(2023)Mitchell, Lee, Khazatsky, Manning, and Finn}]{mitchell2023detectgpt}
Eric Mitchell, Yoonho Lee, Alexander Khazatsky, Christopher~D. Manning, and Chelsea Finn. 2023.
\newblock \href {https://arxiv.org/abs/2301.11305} {Detectgpt: Zero-shot machine-generated text detection using probability curvature}.

\bibitem[{Mitrovi{\'c} et~al.(2023)Mitrovi{\'c}, Andreoletti, and Ayoub}]{mitrovic2023chatgpt}
Sandra Mitrovi{\'c}, Davide Andreoletti, and Omran Ayoub. 2023.
\newblock Chatgpt or human? detect and explain. explaining decisions of machine learning model for detecting short chatgpt-generated text.
\newblock \emph{arXiv preprint arXiv:2301.13852}.

\bibitem[{MosaicML(2023)}]{MosaicML2023Introducing}
NLP~Team MosaicML. 2023.
\newblock \href {https://www.mosaicml.com/blog/mpt-30b} {Introducing mpt-30b: Raising the bar for open-source foundation models}.
\newblock Accessed: 2023-06-22.

\bibitem[{Penedo et~al.(2023)Penedo, Malartic, Hesslow, Cojocaru, Cappelli, Alobeidli, Pannier, Almazrouei, and Launay}]{refinedweb}
Guilherme Penedo, Quentin Malartic, Daniel Hesslow, Ruxandra Cojocaru, Alessandro Cappelli, Hamza Alobeidli, Baptiste Pannier, Ebtesam Almazrouei, and Julien Launay. 2023.
\newblock \href {http://arxiv.org/abs/2306.01116} {The {R}efined{W}eb dataset for {F}alcon {LLM}: outperforming curated corpora with web data, and web data only}.
\newblock \emph{arXiv preprint arXiv:2306.01116}.

\bibitem[{Pu et~al.(2023)Pu, Zhang, Han, Tsvetkov, and He}]{pu2023zero}
Xiao Pu, Jingyu Zhang, Xiaochuang Han, Yulia Tsvetkov, and Tianxing He. 2023.
\newblock On the zero-shot generalization of machine-generated text detectors.
\newblock \emph{arXiv preprint arXiv:2310.05165}.

\bibitem[{Radford et~al.(2019)Radford, Wu, Child, Luan, Amodei, and Sutskever}]{radford2019language}
Alec Radford, Jeff Wu, Rewon Child, David Luan, Dario Amodei, and Ilya Sutskever. 2019.
\newblock Language models are unsupervised multitask learners.

\bibitem[{Sadasivan et~al.(2023)Sadasivan, Kumar, Balasubramanian, Wang, and Feizi}]{sadasivan2023can}
Vinu~Sankar Sadasivan, Aounon Kumar, Sriram Balasubramanian, Wenxiao Wang, and Soheil Feizi. 2023.
\newblock Can ai-generated text be reliably detected?
\newblock \emph{arXiv preprint arXiv:2303.11156}.

\bibitem[{Sarvazyan et~al.(2023{\natexlab{a}})Sarvazyan, Gonz\'{a}lez, Rosso, and Franco-Salvador}]{Sarvazyan2023supervided}
Areg~Mikael Sarvazyan, Jos\'{e}~\'{A}ngel Gonz\'{a}lez, Paolo Rosso, and Marc Franco-Salvador. 2023{\natexlab{a}}.
\newblock \href {https://doi.org/10.1007/978-3-031-42448-9_11} {Supervised machine-generated text detectors: Family and scale matters}.
\newblock In \emph{Experimental IR Meets Multilinguality, Multimodality, and Interaction: 14th International Conference of the CLEF Association, CLEF 2023, Thessaloniki, Greece, September 18–21, 2023, Proceedings}, page 121–132, Berlin, Heidelberg. Springer-Verlag.

\bibitem[{Sarvazyan et~al.(2023{\natexlab{b}})Sarvazyan, Ángel González, Franco-Salvador, Rangel, Chulvi, and Rosso}]{sarvazyan2023overview}
Areg~Mikael Sarvazyan, José Ángel González, Marc Franco-Salvador, Francisco Rangel, Berta Chulvi, and Paolo Rosso. 2023{\natexlab{b}}.
\newblock \href {http://arxiv.org/abs/2309.11285} {Overview of autextification at iberlef 2023: Detection and attribution of machine-generated text in multiple domains}.

\bibitem[{Scao et~al.(2022)Scao, Fan, Akiki, Pavlick, Ili{\'c}, Hesslow, Castagn{\'e}, Luccioni, Yvon, Gall{\'e} et~al.}]{scao2022bloom}
Teven~Le Scao, Angela Fan, Christopher Akiki, Ellie Pavlick, Suzana Ili{\'c}, Daniel Hesslow, Roman Castagn{\'e}, Alexandra~Sasha Luccioni, Fran{\c{c}}ois Yvon, Matthias Gall{\'e}, et~al. 2022.
\newblock Bloom: A 176b-parameter open-access multilingual language model.
\newblock \emph{arXiv preprint arXiv:2211.05100}.

\bibitem[{Solaiman et~al.(2019)Solaiman, Brundage, Clark, Askell, Herbert-Voss, Wu, Radford, Krueger, Kim, Kreps et~al.}]{solaiman2019release}
Irene Solaiman, Miles Brundage, Jack Clark, Amanda Askell, Ariel Herbert-Voss, Jeff Wu, Alec Radford, Gretchen Krueger, Jong~Wook Kim, Sarah Kreps, et~al. 2019.
\newblock Release strategies and the social impacts of language models.
\newblock \emph{arXiv preprint arXiv:1908.09203}.

\bibitem[{Touvron et~al.(2023{\natexlab{a}})Touvron, Lavril, Izacard, Martinet, Lachaux, Lacroix, Rozi{\`e}re, Goyal, Hambro, Azhar et~al.}]{touvron2023llama}
Hugo Touvron, Thibaut Lavril, Gautier Izacard, Xavier Martinet, Marie-Anne Lachaux, Timoth{\'e}e Lacroix, Baptiste Rozi{\`e}re, Naman Goyal, Eric Hambro, Faisal Azhar, et~al. 2023{\natexlab{a}}.
\newblock Llama: Open and efficient foundation language models.
\newblock \emph{arXiv preprint arXiv:2302.13971}.

\bibitem[{Touvron et~al.(2023{\natexlab{b}})Touvron, Martin, Stone, Albert, Almahairi, Babaei, Bashlykov, Batra, Bhargava, Bhosale, Bikel, Blecher, Ferrer, Chen, Cucurull, Esiobu, Fernandes, Fu, Fu, Fuller, Gao, Goswami, Goyal, Hartshorn, Hosseini, Hou, Inan, Kardas, Kerkez, Khabsa, Kloumann, Korenev, Koura, Lachaux, Lavril, Lee, Liskovich, Lu, Mao, Martinet, Mihaylov, Mishra, Molybog, Nie, Poulton, Reizenstein, Rungta, Saladi, Schelten, Silva, Smith, Subramanian, Tan, Tang, Taylor, Williams, Kuan, Xu, Yan, Zarov, Zhang, Fan, Kambadur, Narang, Rodriguez, Stojnic, Edunov, and Scialom}]{touvron2023llama2}
Hugo Touvron, Louis Martin, Kevin Stone, Peter Albert, Amjad Almahairi, Yasmine Babaei, Nikolay Bashlykov, Soumya Batra, Prajjwal Bhargava, Shruti Bhosale, Dan Bikel, Lukas Blecher, Cristian~Canton Ferrer, Moya Chen, Guillem Cucurull, David Esiobu, Jude Fernandes, Jeremy Fu, Wenyin Fu, Brian Fuller, Cynthia Gao, Vedanuj Goswami, Naman Goyal, Anthony Hartshorn, Saghar Hosseini, Rui Hou, Hakan Inan, Marcin Kardas, Viktor Kerkez, Madian Khabsa, Isabel Kloumann, Artem Korenev, Punit~Singh Koura, Marie-Anne Lachaux, Thibaut Lavril, Jenya Lee, Diana Liskovich, Yinghai Lu, Yuning Mao, Xavier Martinet, Todor Mihaylov, Pushkar Mishra, Igor Molybog, Yixin Nie, Andrew Poulton, Jeremy Reizenstein, Rashi Rungta, Kalyan Saladi, Alan Schelten, Ruan Silva, Eric~Michael Smith, Ranjan Subramanian, Xiaoqing~Ellen Tan, Binh Tang, Ross Taylor, Adina Williams, Jian~Xiang Kuan, Puxin Xu, Zheng Yan, Iliyan Zarov, Yuchen Zhang, Angela Fan, Melanie Kambadur, Sharan Narang, Aurelien Rodriguez, Robert Stojnic, Sergey Edunov, and Thomas
  Scialom. 2023{\natexlab{b}}.
\newblock \href {http://arxiv.org/abs/2307.09288} {Llama 2: Open foundation and fine-tuned chat models}.

\bibitem[{Uchendu et~al.(2020)Uchendu, Le, Shu, and Lee}]{uchendu-etal-2020-authorship}
Adaku Uchendu, Thai Le, Kai Shu, and Dongwon Lee. 2020.
\newblock \href {https://doi.org/10.18653/v1/2020.emnlp-main.673} {Authorship attribution for neural text generation}.
\newblock In \emph{Proceedings of the 2020 Conference on Empirical Methods in Natural Language Processing (EMNLP)}, pages 8384--8395, Online. Association for Computational Linguistics.

\bibitem[{Wei et~al.(2022)Wei, Bosma, Zhao, Guu, Yu, Lester, Du, Dai, and Le}]{wei2022finetuned}
Jason Wei, Maarten Bosma, Vincent Zhao, Kelvin Guu, Adams~Wei Yu, Brian Lester, Nan Du, Andrew~M. Dai, and Quoc~V Le. 2022.
\newblock \href {https://openreview.net/forum?id=gEZrGCozdqR} {Finetuned language models are zero-shot learners}.
\newblock In \emph{International Conference on Learning Representations}.

\bibitem[{Zellers et~al.(2019)Zellers, Holtzman, Rashkin, Bisk, Farhadi, Roesner, and Choi}]{zellers2019neuralfakenews}
Rowan Zellers, Ari Holtzman, Hannah Rashkin, Yonatan Bisk, Ali Farhadi, Franziska Roesner, and Yejin Choi. 2019.
\newblock \href {http://papers.nips.cc/paper/9106-defending-against-neural-fake-news.pdf} {Defending against neural fake news}.
\newblock In H.~Wallach, H.~Larochelle, A.~Beygelzimer, F.~d\textquotesingle Alch\'{e}-Buc, E.~Fox, and R.~Garnett, editors, \emph{Advances in Neural Information Processing Systems 32}, pages 9054--9065. Curran Associates, Inc.

\bibitem[{Zhang et~al.(2022)Zhang, Roller, Goyal, Artetxe, Chen, Chen, Dewan, Diab, Li, Lin et~al.}]{zhang2022opt}
Susan Zhang, Stephen Roller, Naman Goyal, Mikel Artetxe, Moya Chen, Shuohui Chen, Christopher Dewan, Mona Diab, Xian Li, Xi~Victoria Lin, et~al. 2022.
\newblock Opt: Open pre-trained transformer language models.
\newblock \emph{arXiv preprint arXiv:2205.01068}.

\bibitem[{Zhao et~al.(2023)Zhao, Zhou, Li, Tang, Wang, Hou, Min, Zhang, Zhang, Dong, Du, Yang, Chen, Chen, Jiang, Ren, Li, Tang, Liu, Liu, Nie, and Wen}]{zhao2023survey}
Wayne~Xin Zhao, Kun Zhou, Junyi Li, Tianyi Tang, Xiaolei Wang, Yupeng Hou, Yingqian Min, Beichen Zhang, Junjie Zhang, Zican Dong, Yifan Du, Chen Yang, Yushuo Chen, Zhipeng Chen, Jinhao Jiang, Ruiyang Ren, Yifan Li, Xinyu Tang, Zikang Liu, Peiyu Liu, Jian-Yun Nie, and Ji-Rong Wen. 2023.
\newblock \href {http://arxiv.org/abs/2303.18223} {A survey of large language models}.

\bibitem[{Zheng et~al.(2023)Zheng, Chiang, Sheng, Zhuang, Wu, Zhuang, Lin, Li, Li, Xing, Zhang, Gonzalez, and Stoica}]{zheng2023judging}
Lianmin Zheng, Wei-Lin Chiang, Ying Sheng, Siyuan Zhuang, Zhanghao Wu, Yonghao Zhuang, Zi~Lin, Zhuohan Li, Dacheng Li, Eric.~P Xing, Hao Zhang, Joseph~E. Gonzalez, and Ion Stoica. 2023.
\newblock \href {http://arxiv.org/abs/2306.05685} {Judging llm-as-a-judge with mt-bench and chatbot arena}.

\bibitem[{Ziegler et~al.(2020)Ziegler, Stiennon, Wu, Brown, Radford, Amodei, Christiano, and Irving}]{ziegler2020finetuning}
Daniel~M. Ziegler, Nisan Stiennon, Jeffrey Wu, Tom~B. Brown, Alec Radford, Dario Amodei, Paul Christiano, and Geoffrey Irving. 2020.
\newblock \href {http://arxiv.org/abs/1909.08593} {Fine-tuning language models from human preferences}.

\end{thebibliography}

\clearpage
\onecolumn
\appendix

\section{Full Cross-Model Detection Results}
\label{appendix:full_table}
\begin{figure*}[h!]
    \centering
    \includegraphics[angle=90,origin=c,width=0.8\columnwidth,height=0.68\textheight]{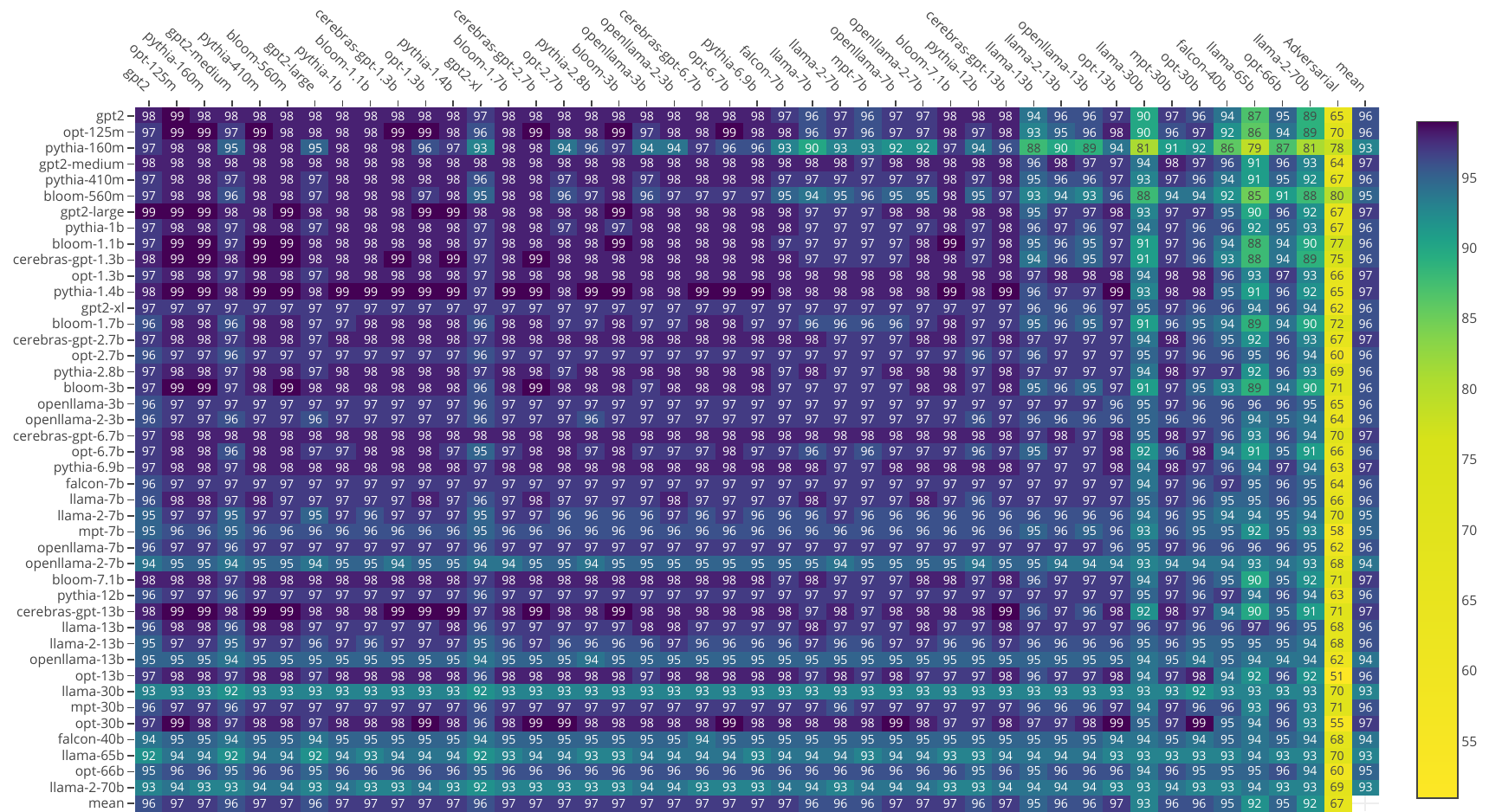}
    \caption{5-seed averaged AUC scores for a classifier trained on text from a source model (\textit{Side axis}) and tested on text from a target model (\textit{Top axis}).}
    \label{fig:res.cross.det.full}
\end{figure*}

\end{document}